\documentclass[letterpaper, 10 pt, conference]{ieeeconf}  

\IEEEoverridecommandlockouts                              

\usepackage{times}
\usepackage{epsfig}
\usepackage{graphicx}
\usepackage{caption}
\usepackage{amsmath}
\usepackage{amssymb}
\usepackage{array,multirow,booktabs}
\usepackage{hyperref}
\usepackage{capt-of,etoolbox}
\usepackage{amsfonts}
\usepackage{hyperref}

\usepackage{cite}
\usepackage{amsmath,amssymb,amsfonts}
\usepackage{algorithmic}
\usepackage{graphicx}
\usepackage{textcomp}
\usepackage[table,xcdraw,dvipsnames]{xcolor}
\usepackage{lipsum}
\usepackage{pifont}

\graphicspath{{./figures/}}
\usepackage{color}
\usepackage[table]{xcolor}
\definecolor{LightCyan}{rgb}{0.88,1,1}
\definecolor{Green}{HTML}{99FF66}

\definecolor{LightCyan}{rgb}{0.88,1,1}
\definecolor{LightYellow}{rgb}{1,1,0.7}
\definecolor{LightGreen}{rgb}{0.4, 1, 0.6}
\newcommand{\xmark}{\ding{55}}

\def\netname{RTS$^2$Net}

\def\eg{\emph{e.g. }}

\def\ie{\emph{i.e. }}
\def\etal{\emph{et al. }}

\begin{document}

\title{Real-Time Semantic Stereo Matching\\
}

\author{Pier Luigi Dovesi$^{1, 2}$, Matteo Poggi$^{3}$, Lorenzo Andraghetti$^1$, Miquel Mart\'i$^{1, 2}$, \\
Hedvig Kjellstr\"{o}m$^{2}$, Alessandro Pieropan$^{1}$,  Stefano Mattoccia$^{3}$
\thanks{$^{1}$Univrses AB}
\thanks{$^{2}$KTH Royal Institute of Technology, Sweden.}
\thanks{$^{3}$University of Bologna, Italy.}
}

\maketitle

\begin{abstract} 
Scene understanding is paramount in robotics, self-navigation, augmented reality,  and many other fields. To fully accomplish this task, an autonomous agent has to infer the 3D structure of the sensed scene (to know where it looks at) and its content (to know what it sees). 
To tackle the two tasks, deep neural networks trained to infer semantic segmentation and depth from stereo images are often the preferred choices. Specifically, Semantic Stereo Matching can be tackled by either standalone models trained for the two tasks independently or joint end-to-end architectures. Nonetheless, as proposed so far, both solutions are inefficient because requiring two forward passes in the former case or due to the complexity of a single network in the latter, although jointly tackling both tasks is usually beneficial in terms of accuracy.
In this paper, we propose a single compact and lightweight architecture for real-time semantic stereo matching.
Our framework relies on coarse-to-fine estimations in a multi-stage fashion, allowing: i) very fast inference even on embedded devices, with marginal drops in accuracy, compared to state-of-the-art networks, ii) trade accuracy for speed, according to the specific application requirements.
Experimental results on high-end GPUs as well as on an embedded Jetson TX2 confirm the superiority of semantic stereo matching compared to standalone tasks and highlight the versatility of our framework on any hardware and for any application.
\end{abstract}

\section{Introduction}

In order to develop a fully autonomous system able to navigate in an unknown environment independently, scene understanding is essential. In particular, an intelligent agent needs to recognize objects in its surroundings and determine their 3D location before performing high-level reasoning concerning path planning, collision avoidance and other tasks. 
This requires addressing two problems: \textit{depth estimation} and \textit{semantic segmentation}. Among the techniques to infer depth, stereo vision has been around for a long time \cite{scharstein2002taxonomy} since it is potentially accurate and efficient. In the few past years it has been heavily influenced by machine learning techniques. In contrast, semantic segmentation only recently emerged as an effectively addressable problem thanks to machine learning and the recent spread of deep learning.  

In this paper, we refer to \textit{Semantic Stereo Matching} as the combination of the two tasks aimed at understanding the surrounding environment sensed by a stereo camera. Nowadays, standalone networks trained for each of the two specific tasks represent the state-of-the-art. However, although modern deep architectures allow for easy integration of multiple tasks \cite{caruana1998Multitaskl}, top performing frameworks rarely exploit the possible synergies between the tasks. 
Indeed, casting semantic stereo matching as a joint optimization of segmentation and disparity estimation yields mutual benefit to both tasks. For instance, depth estimation in challenging portions of the image corresponding to reflective surfaces can be improved by knowing that they belong to a car and thus to an object with defined 3D properties. On the other hand, depth awareness can help to reduce ambiguity when dealing, for instance, with the segmentation of vegetation and terrain.
Several works in the literature support the synergy between semantic and depth inference \cite{ladicky2014pulling,mousavian2016joint,wang2015towards,kendall2018multi,ramirez2018geometry,Tosi_2020_CVPR} and more recently the first semantic stereo matching frameworks appeared \cite{yang2018segstereo,zhang2019dispsegnet}.
However, even if these first attempts confirm the effectiveness of such a paradigm, they are far from real-time performance even on power hungry high-end GPUs. In particular, they barely break the 1 FPS barrier, thus are not ready for deployment in real-world applications.

\begin{figure}
    \centering
    \includegraphics[width=0.485\textwidth]{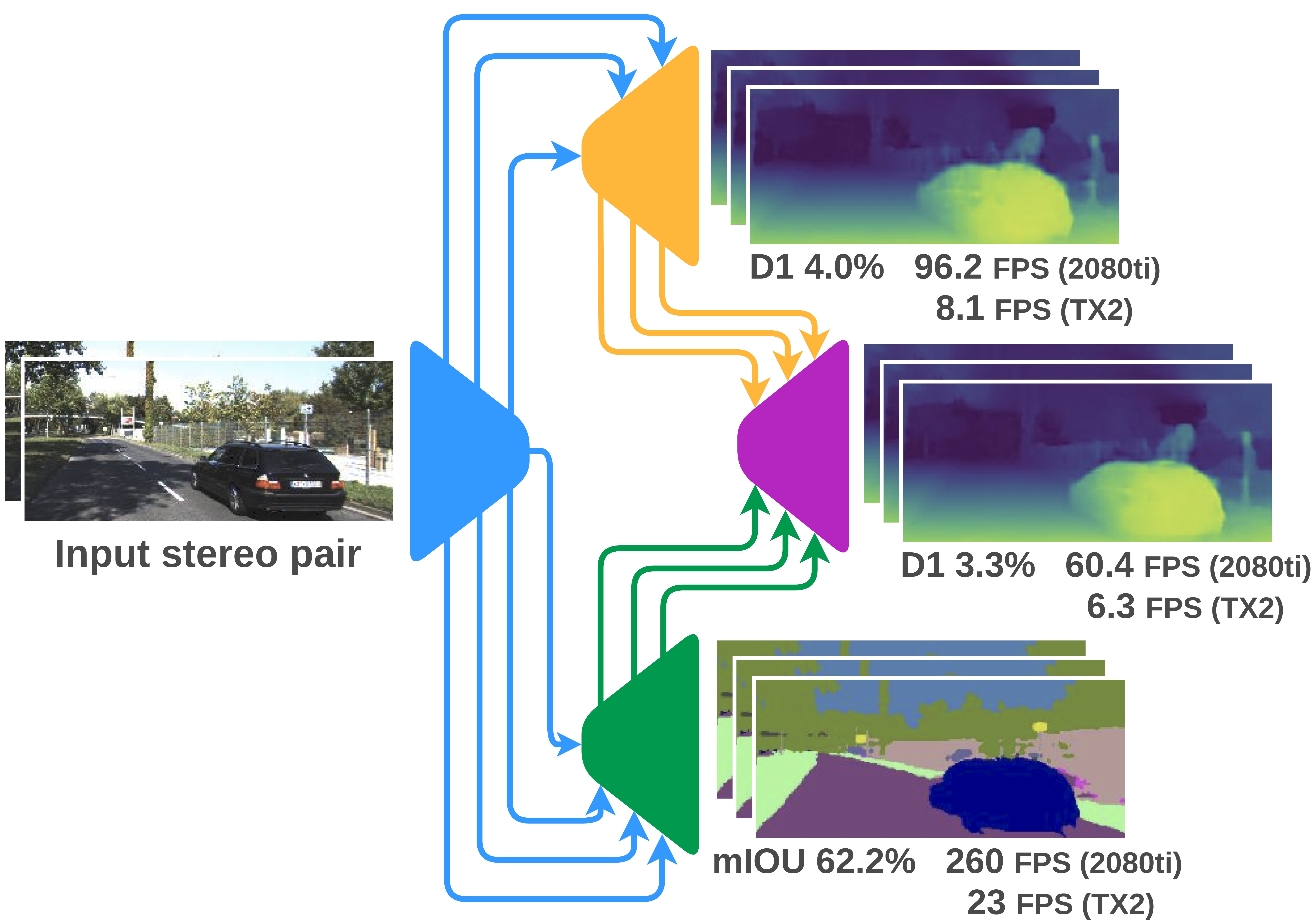}
    \caption{\netname{} allows for fast and accurate semantic segmentation and disparity estimation, both on high-end GPUs and low power systems.}
\label{fig:abstract}
\end{figure}

Purposely, in this paper, we propose a novel \underline{R}eal-\underline{T}ime \underline{S}emantic \underline{S}tereo \underline{Net}work (\netname{}) for jointly solving the two aforementioned tasks. It is designed to leverage the synergies between the two: it learns a common feature representation for both domains and employs separate decoders for estimating accurate semantic segmentation and disparity maps.
Moreover, by designing a stack of multi-stage decoders, \netname{} produces coarse-to-fine estimations for the two tasks, enabling to i) keep low memory and runtime requirements for full inference and ii) further increasing the speed by early-stopping the model at coarse resolution \cite{pydnet18,wang2019anytime} according to the time/resource budget available at deployment.
Figure \ref{fig:abstract} sketches the \netname{} architecture, highlighting how from a shared representation (blue) our network can reason about both semantics (green) and disparity (yellow) and finally post-process early estimates together (purple) to improve depth accuracy. Thanks to its lightweight design, \netname{} can run at several FPS on an NVIDIA Jetson TX2 module with a power budget smaller than 15W, yet providing accurate results competitive with much more complex state-of-the-art networks. Moreover, by early-stopping the network, for instance, before the post-processing phase, we can increase speed with an acceptable decrease of accuracy.
To the best of our knowledge, \netname{} represents the first real-time solution for joint semantic segmentation and stereo matching running seamlessly on high-end GPUs and low-power devices.

\begin{figure*}
    \centering
    \includegraphics[width=0.9\textwidth]{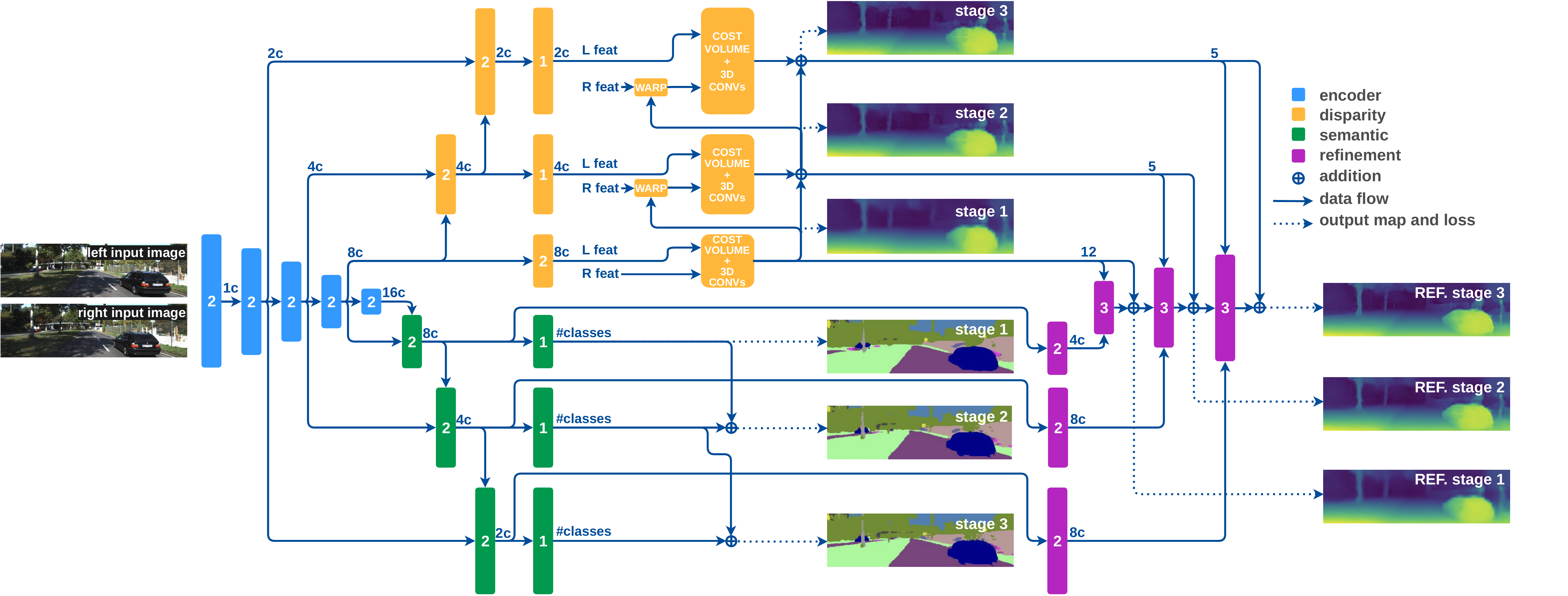}
    \caption{\netname{} architecture overview. Features extracted from the input stereo pair (blue) are the common ground for stereo (yellow) and semantic (green) inference. Finally, the two outputs are combined (purple) for improved synergic disparity estimation. For each block, we report the number of convolutional layers composing it and the number of features they output, multiples of a factor $c$ hyper-parameter of the network.}
    \label{fig:architecture}
\end{figure*}

\section{Related work}

In this section, we review the literature concerning stereo matching, semantic segmentation and multi-task approaches combining depth and semantic.

\textbf{Stereo matching.} Before the deep learning era, stereo algorithms consisted of four well-defined steps \cite{scharstein2002taxonomy}: i) cost computation, ii) cost aggregation, iii) disparity optimization/computation and iv) disparity refinement. 
Eventually, the very first attempts to leverage machine learning for stereo concerned confidence measures \cite{poggi2017quantitative} or replacing some of the aforementioned steps in stereo with deep learning, for example learning a matching function by means of CNNs \cite{zbontar2016stereo,chen2015deep,luo2016efficient}, improving optimization \cite{seki2016patch,seki2017sgm-net} or refining disparity maps \cite{gidaris2017DRR,batsos2018recresnet}. 

End-to-end networks for stereo matching appeared simultaneously to the availability of synthetic data \cite{mayer2016large} and DispNetC was the first network introducing a custom correlation layer to encode similarities between pixels as features. Kendall \etal \cite{kendall2017-gcnet} designed GC-Net, a 3D network processing a cost volume built through features concatenation.
Starting from these seminal works, two families of architectures were developed, respectively 2D and 3D networks.
Frameworks belonging to the first class traditionally use one or multiple correlation layers \cite{pang2017cascade,liang2018learning_1st_Rob,Ilg_2018_ECCV,song2018edgestereo,yang2018segstereo,Tonioni_2019_CVPR,yin2019hierarchical}, while 3D networks build 4D volumes by means of concatenation \cite{chang2018psmnet,Nie_2019_CVPR,tulyakov2018practical,wang2019anytime}, features difference \cite{khamis2018stereonet} or group-wise correlations \cite{guo2019group}.
Although most works focus on accuracy, some deployed compact architectures \cite{Tonioni_2019_CVPR,wang2019anytime,khamis2018stereonet,aleotti2020learning} aimed at real-time performance.
Finally, the guided stereo paradigm \cite{Poggi_CVPR_2019} combines end-to-end models with external depth cues to improve accuracy and generalization of both 2D and 3D architectures.

\textbf{Semantic segmentation.} The advent of deep learning moved semantic segmentation from hand-crafted features and classifiers, like Random Forests \cite{shotton2008semantic} or Support Vector Machines \cite{fulkerson2009class}, to fully convolutional neural networks \cite{long2015fully}.  Architectures for semantic segmentation typically exploit contextual information according to five main strategies. The first consists of using multi-scale prediction models \cite{eigen2015predicting,chen2016attention,chen2018deeplab,liang2015semantic}, making the same architecture process inputs at different scales so to extract features at different contextual levels. The second deploys traditional, encoder-decoder architectures \cite{long2015fully,badrinarayanan2017segnet,ronneberger2015u,lin2016refinenet}. The third encodes long-range context information exploiting Conditional Random Fields either as a post-processing module \cite{chen2018deeplab} or as an integral part of the network \cite{zheng2015conditional}. The fourth uses spatial pyramid pooling to extract context information at different levels \cite{zhao2017pyramid,chen2018deeplab,chen2018deeplab}. Finally, the fifth deploys \textit{atrous}-convolutions to extract higher resolution features while keeping a large receptive field to capture long-range information \cite{dai2017deformable,wang2017understanding}.
As for stereo, some recent works \cite{poudel2018contextnet,poudel2019fast,yu2018bisenet,mazzini2018guided, Chen_2019} focused on efficiency rather than on accuracy for semantic segmentation. Zhu \etal \cite{semantic_cvpr19} recently proposed video prediction-based method to synthesize new training samples.

\textbf{Multi-task frameworks.} There exist approaches aimed at joint depth and semantic estimation, either from monocular images \cite{ladicky2014pulling,mousavian2016joint,wang2015towards,kendall2018multi,ramirez2018geometry,Tosi_2020_CVPR} or stereo images \cite{yang2018segstereo, zhang2019dispsegnet}. In both cases, jointly learning depth and semantic segmentation enabled the improvement of each task. Nonetheless, stereo approaches are lagging far behind the real-time performance required by most practical applications.

\section{Real-time semantic stereo network}

In this section, we introduce our framework for semantic stereo matching.
We start with a general overview of the proposed \netname{}, then focus on describing each component and their interactions.

\subsection{Architecture Overview}
\label{sec:architecture}

In order to achieve high accuracy with limited execution time, the network design consists of a fully residual and pyramidal architecture \cite{pydnet18,wang2019anytime,Tonioni_2019_CVPR}. As depicted in Figure \ref{fig:architecture}, the network is divided into four distinct modules: shared encoder in blue, stereo disparity decoder in yellow, semantic decoder in green and synergy disparity refinement module in purple. For each block, we report the number of convolutional layers composing it and the number of features they output as multiple of a factor $c$, hyper-parameter of the network described in detail next. The network is designed to keep a symmetrical architecture between disparity regression and semantic segmentation in order to facilitate the exploitation of the shared parameters. Both segmentation and disparity are fully computed only at the lowest resolution and progressively refined through the higher resolution residual stages. The same design occurs for the final refinement module, processing the two outputs to improve the disparity estimation significantly. Indeed, even in this final stage, the full refined disparity is only computed at the lowest level and progressively upsampled together with the coarse disparity and semantic segmentation. This fully residual setup provides consistent advantages both at training-time, since early losses stabilize and accelerate this phase, and at testing-time since we can dynamically adjust the speed/accuracy trade-off, as discussed next.

\subsection{Joint features extractor}
As in most architectures, the earliest stage performs feature extraction from the input images. The shared encoder, depicted in blue in Figure \ref{fig:architecture}, is made of two initial $3\times3$ convolutions extracting $c$ features and bringing the resolution to half, then followed by four blocks each one containing a $2\times2$ max-pooling operation and two $3\times3$ layers. The four respectively extract 2$c$, 4$c$, 8$c$, 16$c$ features while progressively halving the resolution, \ie $\frac{1}{4}$, $\frac{1}{8}$, $\frac{1}{16}$ and $\frac{1}{32}$ respectively. Batch normalization and ReLU operations follow all convolutional layers.
Features extracted by this module are processed by two subnetworks, in charge respectively of semantic segmentation and disparity estimation. This forces \netname{} to learn a general and enriched representation meaningful for both tasks.
This design allows us for a dramatic reduction of the computational cost compared to much more complex encoders such as VGG \cite{simonyan2014very}, yet enabling accurate results. In particular, previous works \cite{wang2019anytime} proved that a tiny amount of features, \ie c=1, already enables for decent disparity estimation while significantly increasing the framerate. However, it is insufficient to learn a representation good enough for semantic segmentation too. 

\subsection{Disparity Network}
Following the design of pyramidal networks \cite{wang2019anytime,pydnet18,Tonioni_2019_CVPR}, a stack of decoders is deployed to estimate coarse-to-fine disparity maps.
This strategy allows us to keep computational efforts low as well as to manage the speed-accuracy trade-off dynamically, by performing three stages respectively at $\frac{1}{16}$, $\frac{1}{8}$ and $\frac{1}{4}$ resolution. These stages have been selected because the coarser resolution, \eg at $\frac{1}{32}$, did not improve the results while running decoders at lower-res would significantly increase the runtime with negligible improvements on the final accuracy.
Deploying the shared features computed by the feature extractor, task-specific embeddings are extracted at the three resolutions mentioned above, as shown by the yellow blocks in Figure \ref{fig:architecture}. 

At first, the disparity network takes the disparity features extracted at $\frac{1}{16}$ resolution and builds a distance-based cost volume by progressively shifting right features up to a maximum $d_{max}$ range and subtracting them from left ones to directly obtain an approximation of matching costs. By building the volume at low resolution, a small $d_{max}$ is enough to look for the entire disparity range at the original resolution. In particular, we choose $d_{max}=12$, corresponding to 192 maximum disparity at full resolution.
Then, the volume is regularized through three 3D \textit{conv} blocks followed by batch normalization and ReLU, extracting respectively 16, 16 and 1 features. Finally the disparity map is obtained by means of a \textit{soft-argmin} \cite{kendall2017-gcnet} operator. We kept the same amount of channels as in \cite{wang2019anytime}.
This first, coarse estimation is upsampled to $\frac{1}{16}$ and used to warp right disparity features towards left ones. At this stage, a new cost volume is built in order to find residual disparities and thus to obtain a more accurate disparity map. This time we assume $d_{max}=\pm2$, \ie $\pm16$ at full resolution (we look for both positive and negative residuals, since coarse disparities may be higher or lower than real values).
Then we deploy a decoder with three 3D convolutions extracting 4, 4 and 1 features and a final \textit{soft-argmin} layer as well. The residual disparity is summed to the upsampled estimation from $\frac{1}{16}$ resolution, and the resulting map is further upsampled to $\frac{1}{4}$ resolution for the final stage, identical to the previous, to improve further the disparity estimation.
Finally, the result of the third stage is bilinearly upsampled from $\frac{1}{4}$ to full resolution.

\subsection{Semantic Segmentation Network}
The second subnetwork in charge of semantic segmentation follows the same coarse-to-fine design strategy for the reasons previously outlined as well as to balance the two branches (\ie depth and segmentation) of the whole \netname{} network. 
Again, the shared features computed by the encoder are processed by additional 2D convolutions as in the disparity branch. Besides, $\frac{1}{32}$ features are also used to exploit a broader image context, crucial for semantic segmentation.  
The semantic segmentation branch is made of three stages as well, as shown by the green blocks in Figure \ref{fig:architecture}. Each stage produces per-pixel probability scores for each semantic class, defined according to the KITTI benchmark, at $\frac{1}{16}$, $\frac{1}{8}$ and $\frac{1}{4}$ as the disparity network does. As depicted in the figure, estimated probabilities are upsampled across the stages and summed using residual connections to the outputs of the same stage. These final probabilities allow to infer the semantic map at each stage through a \textit{argmax} over the class scores.

\subsection{Synergy Disparity Refinement module}
The network described so far outputs standalone semantic and disparity maps, yet from a shared representation. The final module in \netname{}, namely \textit{Synergy Disparity Refinement}, reverts this path by jointly processing the two task-specific estimates with a single module to refine the disparity regression leveraging semantic cues. A similar method has been successfully deployed by previous works \cite{yang2018segstereo,zhang2019dispsegnet} with a simple, yet effective strategy consisting of a concatenation of the two embeddings into a \textit{hybrid volume}. 

We adapted this approach to the fully residual strategy followed both in the disparity network and in the semantic decoder. To achieve this, we perform a cascade of residual concatenations between semantic class probabilities and disparity volumes. The refinement module, in purple in Figure \ref{fig:architecture}, performs three steps: 1) in order to limit computational time and balance the contributions in the hybrid volume, we compress the semantic embedding so to have dimensionality similar to the disparity cost volume, 2) we concatenate compressed semantic features with disparity volumes (reorganized so to have disparity dimension as channels) to form the hybrid volumes, in the second and third stage we also concatenate the upsampled previously computed refined disparity, 3) the hybrid volume is then processed through three 2D convolutional layers, producing disparity residuals summed up to the original, reorganized volumes on which the \textit{soft-argmin} operator is applied.

\subsection{Objective function}
Summarizing the network outputs, we have 3 coarse disparities $d_{st}$, 3 semantic segmentation $s_{st}$ and 3 refined disparities $d_{st}^r$, with stages $st \in [1,2,3]$ corresponding to the 3 different resolutions. Regarding the disparity regression, we employ smooth L1 losses $\mathcal{L}_{d_{st}}$ and $\mathcal{L}_{d_{st}^r}$ defined as 
\begin{equation}
L1_{smooth} =
    \begin{cases}
    0.5 (d_i - \hat{d}_i)^2, & \text{if } |d_i - \hat{d}_i| < 1 \\
    |d_i - \hat{d}_i| - 0.5, & \text{otherwise }
    \end{cases}
\end{equation}
with $d$ and $\hat{d}$ respectively the estimated and ground truth disparities,
while $\mathcal{L}_{s_{st}}$ for semantic segmentation \textit{multi class cross entropy}. All losses are averaged over the total amount of pixels.
Since the outputs belong to different decoders and thus computed at different resolutions, we propose a double hierarchical loss weighing scheme:
\begin{equation}
    \mathcal{L} = \sum_{st=1}^{3} \mathcal{W}_{st}\cdot(\mathcal{W}_d\cdot\mathcal{L}_{d_{st}} + \mathcal{W}_s\cdot\mathcal{L}_{s_{st}} + \mathcal{W}_{d^r}\cdot\mathcal{L}_{d_{st}^r})
\end{equation}
where $\mathcal{L}$ is the overall objective function score, $\mathcal{W}_{st}$ are stage weights and $\mathcal{W}_d,\mathcal{W}_s,\mathcal{W}_{d^r}$ are task specific weights respectively for disparity, semantic and refined disparity. In our case $W_{st}$ are respectively $\frac{1}{4}$, $\frac{1}{2}$ and 1 for first, second and final stages, while $\mathcal{W}_d$, $\mathcal{W}_s$, $\mathcal{W}_{d^r}$ are 1, 2 and 2.
The segmentation cross-entropy is also weighted according to the class probability to alleviate the effect of unbalanced datasets \cite{paszke2016enet}. Moreover, since we are working under a multi-task setup, we want to keep the impact of the segmentation independent to the choice of internal weighing schedule or class distribution. Therefore, we design the following weighing scheme:
\begin{equation}
\label{eq:class_weights}
\mathcal{W}_{j}= \frac{N}{\log \left(\mathcal{P}_{j}+k\right)\sum_{i=1}^{N} \frac{1}{\log \left(\mathcal{P}_{i}+k\right)}}
\end{equation}
with $W_j$ the weight of the $j$ class, $N$ the total number of classes, $P$ a class probability and $k$ a parameter that controls the variance of the class weights, set differently according to the dataset (\ie, 1.12 for CityScapes \cite{Chen_2019} and 2 for KITTI 2015). Finally, in case of coarse semantic annotations \cite{cordts2016cityscapes}, we re-weight the segmentation loss according to the percentage of unlabelled area left in the ground truth to obtain $\mathcal{L}_{s}*$

\begin{equation}
    \mathcal{L}_{s}* = \mathcal{L}_{s} (1 + \gamma \cdot \frac{A_{unlab}}{A_{tot} - A_{unlab}})
\end{equation}
with $\gamma$ set to 0.1, to achieve the best results, and $A_{unlab}$, $A_{tot}$ respectively the unlabelled and total amounts of pixels.

\section{Experimental Results}
In this section, we extensively evaluate the performance of \netname{} in terms of both accuracy and runtime. To compare different variants of our model and measure the impact of each of the design choices, we report quantitative results on a validation split sampled from the KITTI 2015 training split made of 40 images, using the remaining 160 for training.
We report the End-Point-Error (EPE) and percentage of pixels with disparity error larger than 3 pixels and 5\% of the ground truth (D1-all\%) to evaluate the accuracy of estimated disparity maps. For both metrics: the lower, the better. For semantic segmentation, we compute the class mean Intersection Over Union (mIOU\%) and the per-pixel accuracy (pAcc\%). For both metrics: the higher, the better.

\begin{table}
\centering
\resizebox{0.45\textwidth}{!}
{
\begin{tabular}{|c|c|c|cc|}
\multicolumn{3}{c}{} &\multicolumn{2}{c}{\cellcolor{blue!25} Disparity }  \\
\hline
Main dataset & epochs & KITTI epochs & \cellcolor{blue!25} EPE & \cellcolor{blue!25} D1-all\% \\

\hline
Sceneflow & 10 & 300 & 1.24 & 6.47 \\
Sceneflow & 40 & 800 & 1.18 & 6.28 \\
CS (coarse$\xrightarrow{}$fine) & 60$\xrightarrow{}$75 & 800 & \textbf{1.14} & \textbf{5.75} \\
\hline
\end{tabular}
}
\caption{Different training schedules / train sets tested on KITTI 2015 validation split, with c=1 (AnyNet \cite{wang2019anytime}). }
\label{table:schedule}
\end{table}

\subsection{Training Schedule}

Traditionally, end-to-end stereo networks are trained from scratch on the Freiburg SceneFlow dataset \cite{mayer2016large}, an extensive collection of synthetic stereo images with dense ground truth disparities, before finetuning on the real, yet smaller target dataset such as KITTI 2015 \cite{menze2015object}. However, since SceneFlow does only provide instance segmentation labels, it is not possible to train \netname{} for semantic segmentation on such imagery. Thus, we initialize our network on the CityScapes dataset\cite{cordts2016cityscapes} (CS), providing about 25K stereo pairs with disparity maps obtained employing Semi-Global Matching algorithm (SGM) \cite{hirschmuller08} and semantic segmentation labels, for which 5K images are densely labeled and 20K coarsely. Although disparity ground truth maps are noisy, a proper training schedule on CS in place of the traditional SceneFlow dataset is more effective when moving to KITTI.
Table \ref{table:schedule} reports experiments supporting this strategy. We trained a variant of \netname{} by setting c=1 and removing both semantic and refinement networks, \ie equivalent to the AnyNet architecture \cite{wang2019anytime}. This way, we aim at measuring only the impact of the different training schedules on disparity estimation, excluding improvements introduced by model variants or multi-task learning that will be evaluated in the remainder. In all our experiments, we train on $256\times512$ crops with batch size 6. We use Adam as optimizer with betas 0.9 and 0.999 and set learning rate to 5e$^{-4}$, kept constant on SceneFlow/CityScapes and halved every 200 epochs on KITTI.
We can see how a more extended training on both SceneFlow and KITTI is beneficial compared to the scheduling proposed in \cite{wang2019anytime}, respectively extending from 10 to 40 and from 300 to 800 epochs. By replacing the SceneFlow pre-train with a multi-stage schedule on CS, 60 epochs on coarse ground truth followed by 75 on fine annotations, allows for better accuracy when followed by the same KITTI finetuning.

\begin{table}[t]
\centering
\resizebox{0.45\textwidth}{!}
{
\begin{tabular}{|c|c|cc|cc|cc|}
\multicolumn{2}{c}{} &\multicolumn{2}{c}{\cellcolor{blue!25} Disparity} & \multicolumn{2}{c}{\cellcolor{orange!50} Semantic} & \multicolumn{2}{c}{\cellcolor{green!50} Frame rate (FPS)} \\
\hline
Model & $c$ & {\cellcolor{blue!25} EPE} & {\cellcolor{blue!25} D1-all\%} & \cellcolor{orange!50} mIOU\% &  \cellcolor{orange!50} pAcc\% &{\cellcolor{green!50} TX2} & {\cellcolor{green!50} 2080ti} \\

\hline
AnyNet \cite{wang2019anytime} & 1 & 1.14 & 5.75 & \xmark & \xmark & \bfseries 10.4 & \textbf{96.8}\\
\netname{} & 1 & \bfseries 1.12 & \bfseries 5.57 & 58.86 & 80.86 & 8.3 & 60.5\\

\hline
AnyNet \cite{wang2019anytime} & 4 & 0.96 & 4.22 & \xmark & \xmark & \bfseries 9.3 & \textbf{96.2}\\
\netname{} & 4 & \bfseries 0.90 & \bfseries 3.80 & 60.93 & 89.77 & 7.4 & 60.5 \\
\hline
AnyNet \cite{wang2019anytime} & 8 & 0.91 & 3.98 & \xmark & \xmark & \bfseries 8.1 & \textbf{96.2}\\
\netname{} & 8 & \bfseries 0.84 & \bfseries 3.33 & 62.22 & 90.64 & 6.3 & 60.4 \\
\hline
AnyNet \cite{wang2019anytime} & 16 & 0.87 & 3.52 & \xmark & \xmark & \bfseries 6.2 & \textbf{95.8}\\
\netname{} & 16 & \bfseries 0.78 & \bfseries 2.90 & 67.41 & 92.92 & 4.5 & 60.4 \\
\hline
AnyNet \cite{wang2019anytime} & 32 & 0.82 & 3.12 & \xmark & \xmark & \bfseries 3.5 & \textbf{64.1}\\
\netname{} & 32 & \bfseries 0.74 & \bfseries 2.62 & \bfseries 69.62 & \bfseries 93.57 & 2.3 & 42.2 \\
\hline

\end{tabular}
}
\caption{Impact of $c$ on KITTI 2015 validation split.} 
\label{table:channels}
\end{table}

\subsection{Model variants}

As described in Section \ref{sec:architecture}, we designed most layers in \netname{} to extract features that are multiples of a basis factor $c$. For instance, by cutting off semantic and synergy modules and setting $c$ to 1, we obtain the AnyNet architecture \cite{wang2019anytime}. Although very fast for disparity inference alone, extracting so few features may lack at representing semantical information.
To assess this, we train and evaluate variants of \netname{} by setting different $c$ factors. Table \ref{table:channels} collects the outcome of these experiments conducted on both the AnyNet architecture, inferring disparity only, and our proposal, inferring disparity and semantic segmentation. All networks have been trained following the best schedule discussed in the previous section, \ie 60 epochs on coarse CS, 75 on fine CS and 800 on KITTI.

By setting c=1, we obtain the same AnyNet configuration reported in \cite{wang2019anytime}. Choosing the same $c$ on \netname{} allows for a moderate improvement on disparity estimation, as well as to obtain reasonable results in terms of semantic segmentation, at the cost of lower frame rate.
By increasing $c$ respectively to 4, 8, 16 and 32 we observe a consequent increase in accuracy on both tasks. In particular, passing from 1 to 32 allows for a significant improvement regarding semantic segmentation estimates, confirming that c=1 is insufficient for this purpose.
Interestingly, the margin between \netname{} and AnyNet on disparity metrics gets larger by increasing the number of features. Indeed, EPE margin is 0.02, 0.06, 0.07, 0.09 and 0.12, while D1-all\% margin is 0.18, 0.42, 0.65, 0.62 and 0.50. This highlights that increasing the number of features is much more beneficial when the model is trained jointly for semantic segmentation too, confirming this latter task to benefit more from a larger pool of features.

For practical applications, c=8 represents a good trade-off allowing for 6.3 FPS on Jetson TX2, \ie about 160ms per inference. Most of the time is taken by the disparity subnetwork (120ms).

\begin{table}[t]
\centering
\resizebox{0.48\textwidth}{!}
{
\begin{tabular}{|l|ll|cc|cc|}
\multicolumn{1}{c}{} &\multicolumn{2}{c}{\cellcolor{blue!25} Disparity } & \multicolumn{2}{c}{\cellcolor{orange!50} Semantic } & \multicolumn{2}{c}{\cellcolor{green!50} Frame rate (FPS)} \\
\hline
Networks & \cellcolor{blue!25} EPE & \cellcolor{blue!25} D1-all\% & \cellcolor{orange!50} mIOU\% &  \cellcolor{orange!50} pAcc\% & {\cellcolor{green!50} TX2} & {\cellcolor{green!50} 2080ti} \\
\hline
Disp. & 0.91 & 3.98 & \xmark & \xmark & \bfseries 8.1 & \textbf{96.2}\\
Disp. + Sem. & 0.90 & 3.90 & \textbf{64.21} & \textbf{91.56} & 6.6 & 76.9 \\
Disp. + Sem. + Ref. & 0.91 (\textbf{0.84})  & 3.91 (\textbf{3.33}) & 62.22 & 90.64 & 6.3 & 60.4 \\
\hline 
\end{tabular}
}
\smallskip
\caption{Ablation study (c=8), KITTI 2015 validation split.}
\label{table:multitask}
\end{table}

\begin{table}[t]
\centering
\resizebox{0.48\textwidth}{!}
{
\begin{tabular}{|c|cc|cc|cc|}
\multicolumn{1}{c}{} &\multicolumn{2}{c}{Stage 1} & \multicolumn{2}{c}{Stage 2} & \multicolumn{2}{c}{Stage 3} \\
\hline
Model & {\cellcolor{green!50} FPS} & {\cellcolor{blue!25} D1-all\%} & {\cellcolor{green!50} FPS} & {\cellcolor{blue!25} D1-all\%} & {\cellcolor{green!50} FPS} & {\cellcolor{blue!25} D1-all\%} \\

\hline
AnyNet \cite{wang2019anytime}  & 34.6 & 11.60 & 20.5 & 8.40 & \textcolor{red}{10.4} & \textcolor{red}{5.75} \\
\netname{} (c=8) & 17.2 & 8.00 & \textcolor{red}{10.9} & \textcolor{red}{4.70} & 6.3 & 3.33 \\
\hline
\end{tabular}
}
\smallskip
\caption{Anytime inference, KITTI 2015 validation split.} 
\label{table:anytime}
\end{table}

\begin{figure*}
    \centering
    \renewcommand{\tabcolsep}{1pt}
    \begin{tabular}{ccccc}
    \includegraphics[width=0.18\textwidth]{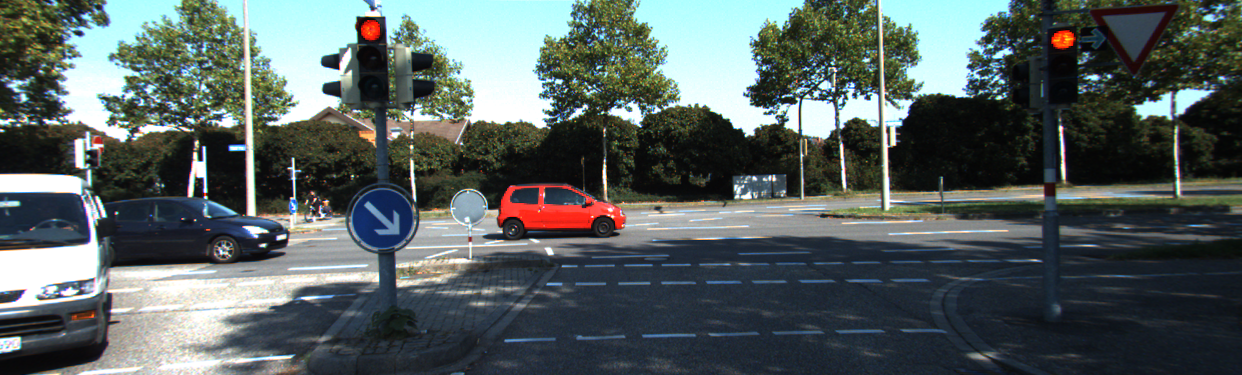} &
    \includegraphics[width=0.18\textwidth]{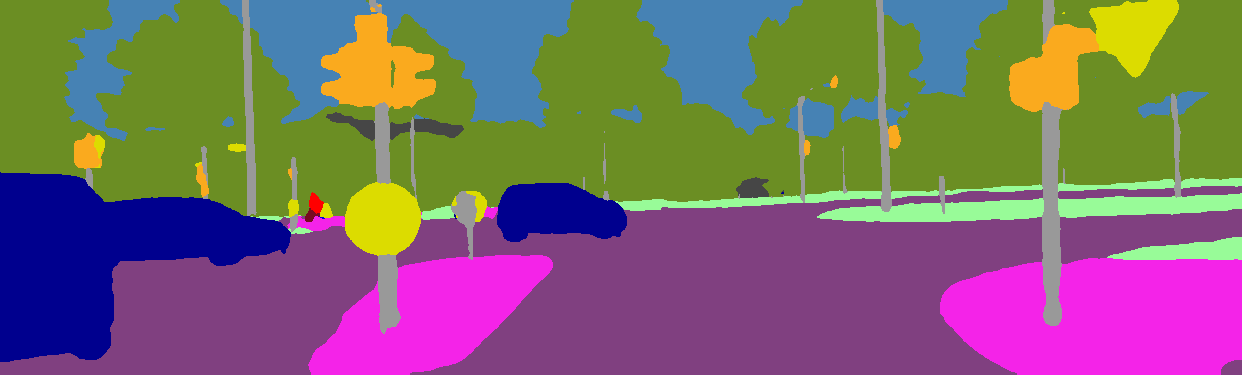} &    
    \includegraphics[width=0.18\textwidth]{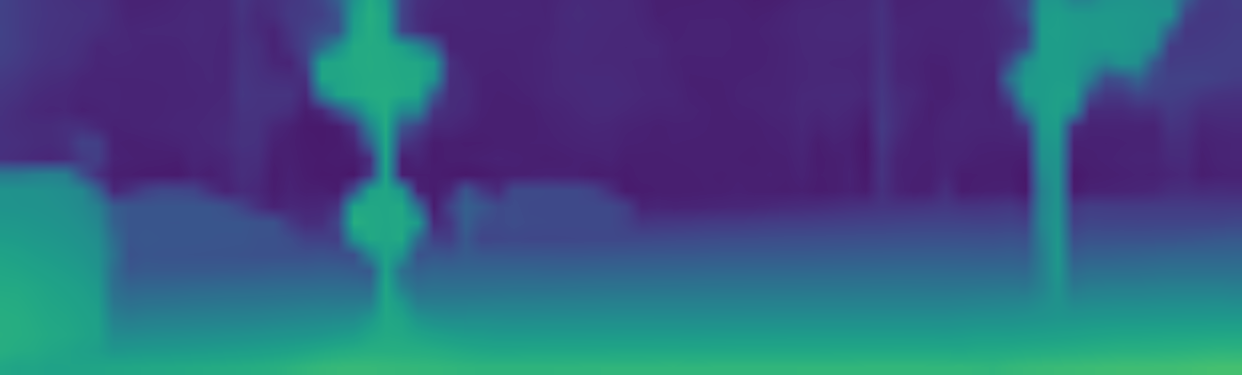} &
    \includegraphics[width=0.18\textwidth]{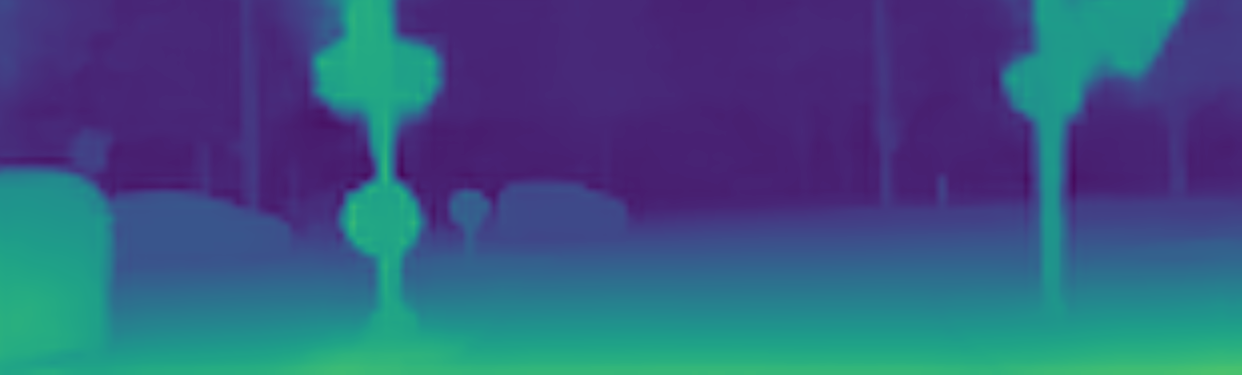} &
    \includegraphics[width=0.18\textwidth]{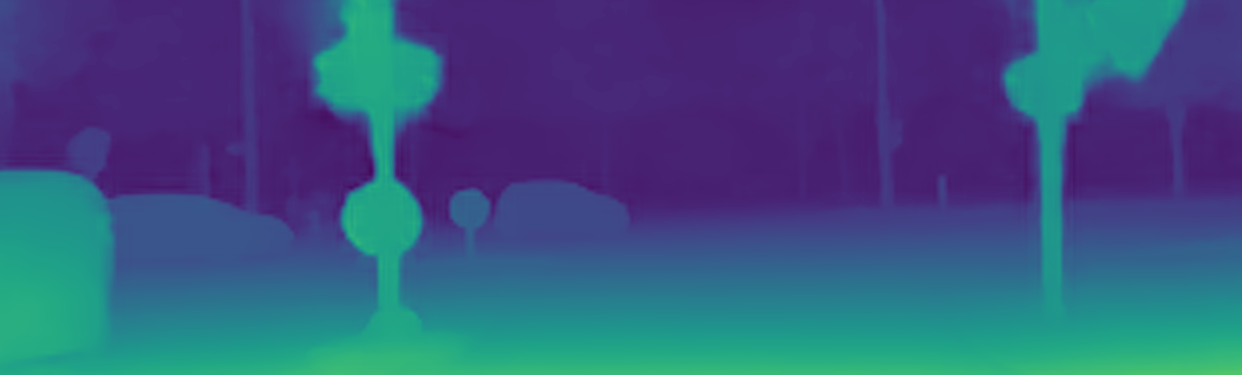} \\
    \includegraphics[width=0.18\textwidth]{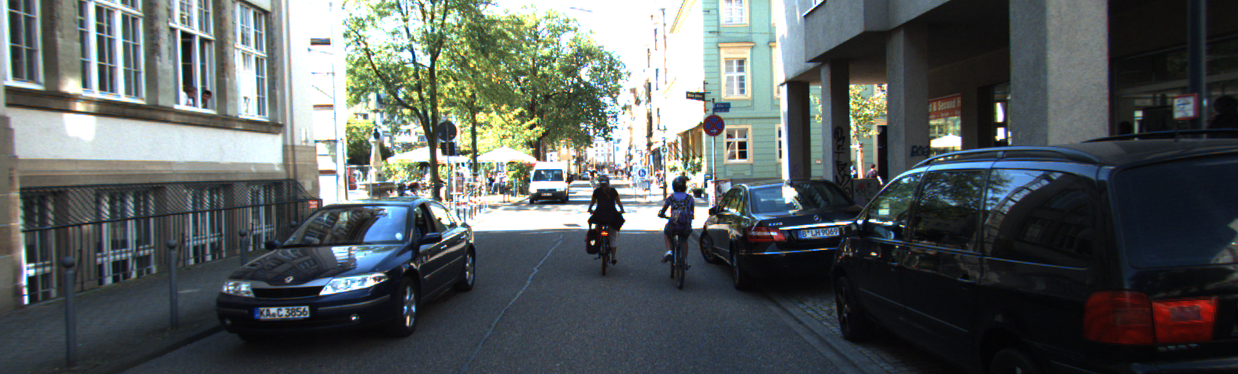} &
    \includegraphics[width=0.18\textwidth]{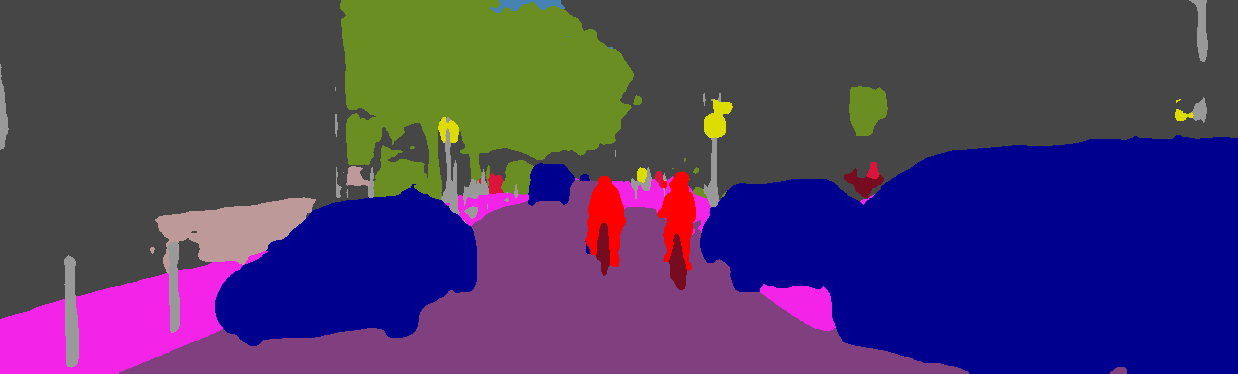} &    
    \includegraphics[width=0.18\textwidth]{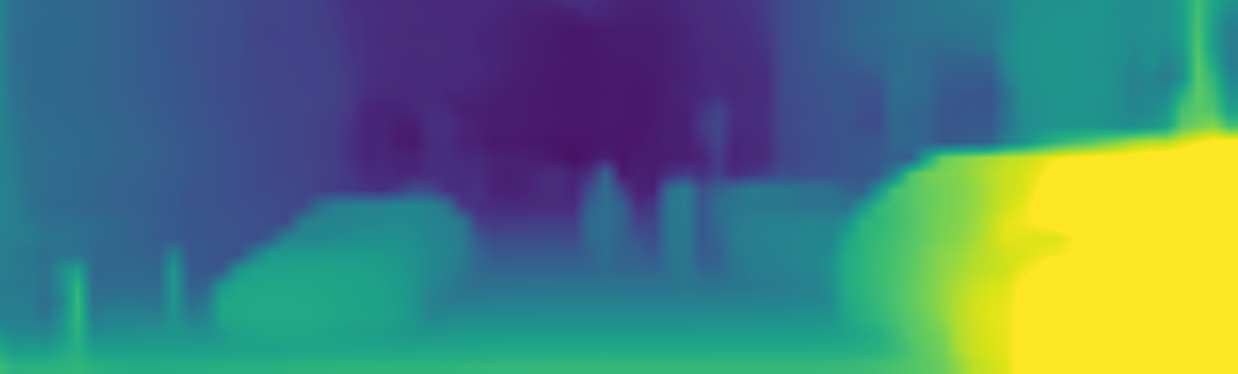} &
    \includegraphics[width=0.18\textwidth]{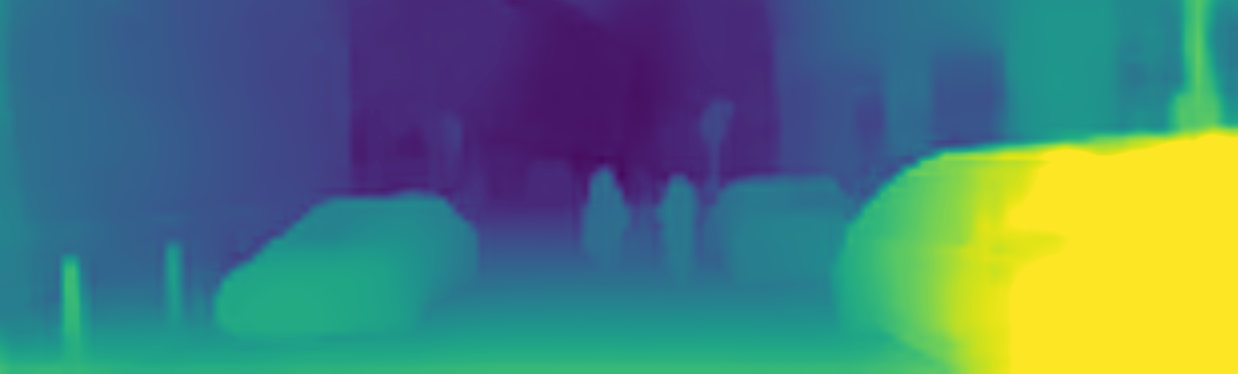} &
    \includegraphics[width=0.18\textwidth]{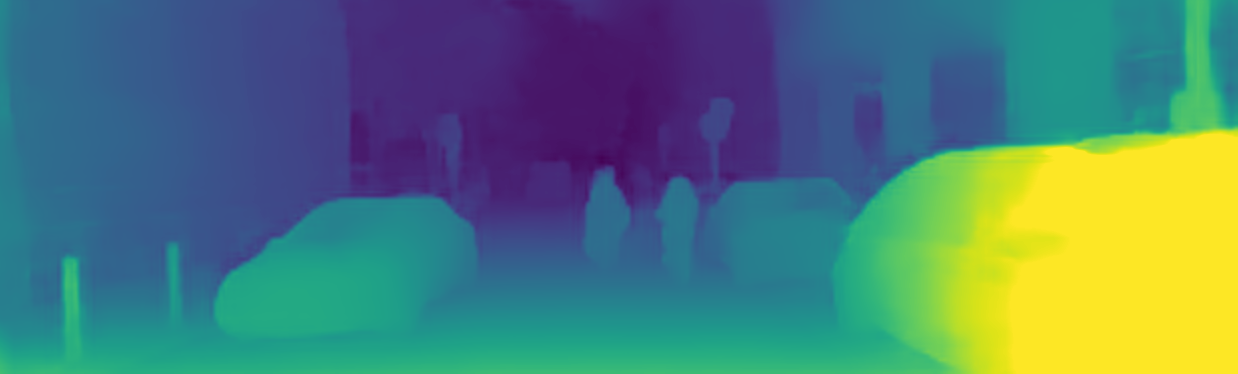} \\
    \end{tabular}
    \caption{Qualitative results on KITTI. From left to right: reference image, semantic and coarse to fine disparity maps.}
\label{fig:qualitatives}
\end{figure*}

\subsection{Impact of multi-task and synergy modules}

We measure the contribution given by both the multi-task learning paradigm itself and the synergy refinement module specifically designed for \netname{}. Table \ref{table:multitask} collects the results obtained by training ablated configuration of our model depicted in Figure \ref{fig:architecture}, setting c=8. The training has been conducted on CS and KITTI, as described in the previous section. On top, the variant made of the features encoder (blue) and the disparity subnetwork (yellow). We can notice how, by simply adding the semantic network (green) and training for joint optimization of the two tasks, slightly increases the disparity accuracy respectively by 0.01 and 0.08 in terms of EPE and D1-all\%. As expected, the best results are achieved by adding the synergy refinement module (purple), shown in brackets on the last row of the table together with EPE and D1-all\% obtained from the disparity network without applying the refinement. From these results, as for semantic segmentation, we can notice that depth estimation marginally loses accuracy compared to the previous model (0.01 on both EPE and D1-all\% and 1.99, 0.92 on mIOU\% and pAcc\%). Nonetheless, this configuration yields a more considerable improvement after refinement.

\subsection{Anytime inference}

\netname{} allows for trading accuracy for speed by early-stopping inference at an intermediate stage, a property shared with other architectures \cite{pydnet18,wang2019anytime}.
Table \ref{table:anytime} compares the trade-off achieved respectively by AnyNet \cite{wang2019anytime} and our architecture measured on the NVIDIA Jetson TX2. We focus on studying the impact on disparity estimation, since it represents the bottleneck in our system.
First, we can notice how \netname{} at any stage runs roughly at half the frames per second, with ample margins in terms of improved accuracy. Moreover, we highlight in red the two configurations achieving the minimum frame rate compatible with the KITTI acquisition system (\ie, 10 FPS \cite{Geiger2012CVPR}), respectively AnyNet Stage 3 and \netname{} Stage 2. In this setting, \netname{} runs slightly faster than AnyNet and achieves 1.05\% reduction in terms of D1-all\%, yet providing the additional semantic segmentation output making our framework the preferred choice for practical applications. 
Moreover, by paying a reasonable price in terms of speed \netname{} can further reduce the error rate compared to AnyNet by a total 2.42\%.

\begin{table}[t]
\centering
\resizebox{0.45\textwidth}{!}
{
\begin{tabular}{|l|ccc|c|}
\hline
Network & \cellcolor{blue!25} D1-bg\% & \cellcolor{blue!25} D1-fg\% & \cellcolor{blue!25} D1-all\% & {\cellcolor{green!50} Runtime (s)} \\
\hline
GANet \cite{zhang2019ga} & 1.48 & 3.46 & 1.81 & 1.80 \\
HD$^3$ \cite{yin2019hierarchical} & 1.70 & 3.63 & 2.02 & 0.14 \\
GWCNet \cite{guo2019group} & 1.74 &	3.93 &	2.11 & 0.32 \\
SegStereo \cite{yang2018segstereo} & 1.88 &	4.07 & 2.25 & 0.60 \\
PSMNet \cite{chang2018psmnet} & 1.86 & 4.62 & 2.32 & 0.41 \\	
\hline
\netname{} (ours) & 3.09 & 5.91 & 3.56 & 0.02 \\
DispNetC \cite{mayer2016large} & 4.32 & 4.41 & 4.34 & 0.06 \\
MADNet \cite{Tonioni_2019_CVPR} & 3.75 & 9.20 &	4.66 & 0.02 \\
StereoNet \cite{khamis2018stereonet} & 4.30 & 7.45 & 4.83 & 0.02 \\
\hline 
\end{tabular}
}
\smallskip
\caption{Result on KITTI 2015 online benchmark (stereo).}
\label{table:online-stereo}
\end{table}

\subsection{Evaluation on KITTI online benchmark}

We report the results achieved by submitting the maps produced by \netname{} on KITTI 2015 online benchmark. To this aim, we trained a model having c=32 to compete with state-of-the-art architectures, traditionally more complex, achieving 0.74, 2.62 in terms of EPE and D1-all\% and  69.62, 93.57 on mIOU\% and pAcc\% on the validation split of Table \ref{table:channels}. We report runtimes on NVIDIA 2080ti.

Table \ref{table:online-stereo} report a comparison between our model and published state-of-the-art architectures taken from the online stereo leaderboard, reporting the D1 metric on the background (D1-bg\%), foreground (D1-fg\%) and all (D1-all\%) pixels. Unfortunately, results for AnyNet were not submitted by the authors to the online KITTI leaderboard. Nonetheless, previous experimental results highlighted the superior accuracy of our proposal. 
From the table, we can notice how \netname{} results more accurate than state-of-the-art real-time frameworks MADNet \cite{Tonioni_2019_CVPR} and StereoNet \cite{khamis2018stereonet}, confirming the effectiveness of jointly inferring semantic and disparity estimation. The gap with state-of-the-art architectures reported in the upper portion of Table \ref{table:online-stereo} ranges between 1.2 and 1.7\% on D1-all\%, yet running 7 to 90$\times$ faster.

\begin{table}[t]
\centering
\resizebox{0.45\textwidth}{!}
{
\begin{tabular}{|l|cccc|c|}
\hline
& \cellcolor{orange!50} IoU & \cellcolor{orange!50} iIoU & \cellcolor{orange!50} IoU & \cellcolor{orange!50} iIoU & {\cellcolor{green!50} Runtime } \\
Network & \cellcolor{orange!50} class\% & \cellcolor{orange!50} class\% & \cellcolor{orange!50} category\% & \cellcolor{orange!50} category\% & {\cellcolor{green!50} (s)} \\
\hline
VideoProp-LabelRelax \cite{semantic_cvpr19} & 72.82 & 48.68 & 88.99 & 75.26 & - \\
IfN-DomAdap-Seg	\cite{Bolte_2019_CVPR_Workshops} & 59.50 & 30.28 & 81.57 & 61.91 & 1.00 \\
\cellcolor{yellow!50} SegStereo \cite{yang2018segstereo} & \cellcolor{yellow!50}59.10 & \cellcolor{yellow!50}28.00 & \cellcolor{yellow!50}81.31 & \cellcolor{yellow!50}60.26 & \cellcolor{yellow!50}0.60 \\
\cellcolor{yellow!50} \netname{} (ours) & \cellcolor{yellow!50}57.67 & \cellcolor{yellow!50}27.42 & \cellcolor{yellow!50}82.85 & \cellcolor{yellow!50}60.72 & \cellcolor{yellow!50}0.02 (0.008)\\
SDNet \cite{OchsKretzMester2019} & 51.14 & 17.74 & 79.62 & 50.45 & 0.20 \\
APMoE\_seg\_ROB \cite{kong2019pag} & 47.96 & 17.86 & 78.11 & 49.17 & 0.20 \\
\hline 
\end{tabular}
}
\smallskip
\caption{Result on KITTI 2015 segmentation benchmark.}
\label{table:online-segmentation}
\end{table}

Table \ref{table:online-segmentation} reports a comparison between \netname{} and published methods on the KITTI semantic segmentation online benchmark, highlighting semantic stereo frameworks in yellow. Regarding the execution time of \netname{}, we report it when regressing only semantic information (0.008s) and depth plus semantic (0.02s). Compared to SegStereo \cite{yang2018segstereo}, our network performs slightly worse on class level, while being more accurate on categories and running about 30$\times$ faster. Moreover, it outperforms some competitors specifically trained for semantic segmentation only \cite{OchsKretzMester2019,kong2019pag}.

\subsection{Qualitative results }

Figure \ref{fig:qualitatives} shows some qualitative examples of semantic segmentation and disparity maps generated by \netname{}.
Finally, we refer the reader to the \textbf{supplementary material}\footnote{\url{https://www.youtube.com/watch?v=wbtQcWAbgo0}} for qualitative results on a KITTI video sequence.

\section{Conclusions}

This paper proposed a fast and lightweight end-to-end deep network for scene understanding capable of jointly inferring depth and semantic segmentation exploiting their synergy. As reported in the exhaustive experimental results, this strategy compares favorably to the state-of-the-art in both tasks. Moreover, a peculiar pyramidal design strategy enables us to infer stereo and semantic segmentation in a fraction of the time required by other methods as well as to dynamically trade accuracy for speed according to the specific application requirements. To the best of our knowledge, our proposal is the first network enabling to simultaneously infer accurate depth and semantic segmentation suited for real-time applications, even on a low power budget deploying embedded devices like the NVIDIA Jetson TX2.

\bibliographystyle{IEEEtran}
\bibliography{ref}

\end{document}